\documentclass[11pt,a4paper]{article}
\usepackage[hyperref]{ranlp2023}
\usepackage{times}
\usepackage{latexsym}

\usepackage{microtype}

\aclfinalcopy % Uncomment this line for the final submission
\def\aclpaperid{96} %  Enter the acl Paper ID here

\usepackage{graphicx} 
\usepackage{tabularx}
\usepackage{comment}
\usepackage{algorithm}
\usepackage{booktabs}
\usepackage{multirow}
\usepackage{array}

\usepackage[noend]{algpseudocode}

\title{\textsc{WikiTiDe}: A \textsc{Wikipedia-based Timestamped Definition} Pairs Dataset}

\author{Hsuvas Borkakoty$^\ast$, Luis Espinosa-Anke$^\ast$\textsuperscript{$\diamond$} \\
$^\ast$Cardiff NLP, School of Computer Science and Informatics, Cardiff University, UK \\
\textsuperscript{$\diamond$}AMPLYFI, UK\\
  \texttt{\{borkakotyh,espinosaankel\}@cardiff.ac.uk} \\}

\date{}

\begin{document}
\maketitle
\begin{abstract}
A fundamental challenge in the current NLP context, dominated by language models, comes from the inflexibility of current architectures to ``learn'' new information. While model-centric solutions like continual learning or parameter-efficient fine-tuning are available, the question still remains of  how to reliably identify changes in language or in the world. In this paper, we propose WikiTiDe, a dataset derived from pairs of timestamped definitions extracted from Wikipedia. We argue that such resource can be helpful for accelerating diachronic NLP, specifically, for training models able to scan knowledge resources for core updates concerning a concept, an event, or a named entity. Our proposed end-to-end method is fully automatic, and leverages a bootstrapping algorithm for gradually creating a high-quality dataset. Our results suggest that bootstrapping the seed version of WikiTiDe leads to better fine-tuned models. We also leverage fine-tuned models in a number of downstream tasks, showing promising results with respect to competitive baselines\footnote{\url{https://github.com/hsuvas/wikitide}}.

%Information is dynamic, it gets updated with time. Detecting whether there is a change in the information is important in the current scenario, as it can help in the efficient updation of the knowledge base. We pose this as a classification problem, where we train a classifier using weak supervision to find if there are any changes in information with respect to two timestamps. We create a small seed dataset for this purpose, containing 3000 timestamped definition pairs extracted from Wikipedia, and train our preliminary model on this dataset for  three classes, to signify whether the change is informational, cosmetic, or no change at all. We construct a big unlabelled dataset to move forward with a weak supervision paradigm, where the model is iteratively updated by bootstrapping the training process with new data from the unlabelled dataset. We find that by considering this approach, we start getting better results as new data is added. We also find that a weak supervision approach helps to create a more robust classifier, able to classify sets of definitions into one of three classes.
\end{abstract}

\section{Introduction}

Handling new information is one of the most critical (and vastly unresolved) challenges in the current NLP landscape, mostly because language models (LMs) such as BERT \cite{devlin-etal-2019-bert}, T5 \cite{raffel2020exploring}, GPT-3 \cite{brown2020language} or PaLM \cite{chowdhery2022palm} can only learn from information they have seen during pretraining. This is an important limitation when it comes to dealing with updates in the world and language changes alike, since these updates, if not dealt with properly in an LM-centric system, can cause \textit{temporal misalignment} \cite{luu2021time,lazaridou2021mind,jang2022temporalwiki}, which is especially harming in knowledge-intensive tasks, such as closed-book QA. 

\begin{table*}[!t]
\centering
%\resizebox{\textwidth}{!}{%
\begin{tabular}{|p{1.5cm}|p{11cm}|m{0.5cm}|}
\toprule
\multicolumn{1}{c}{} & \multicolumn{1}{|c|}{WikiTiDe Definitions} & \multicolumn{1}{r}{Label} \\ \midrule
$p_{first}^{def}$ & "All or Nothing" is a song by German dance-pop group Milli Vanilli. & \multirow{2}{*}{0} \\ \cline{2-2}
$p_{second}^{def}$ & "All or Nothing" is a song by German dance-pop group Milli Vanilli. & \\ \midrule
$p_{first}^{def}$ & "Along the Navajo Trail" is a country/pop song, written by Dick Charles (pseudonym for Richard Charles Krieg), Larry Markes, and Edgar De Lange in 1945. & \multirow{6}{*}{1} \\ \cline{2-2}
$p_{second}^{def}$ & "Along the Navajo Trail" is a country/pop song, written by Dick Charles (pseudonym for Richard Charles Krieg), Larry Markes, and Eddie DeLange in 1945. & \\ \midrule
$p_{first}^{def}$ & Alan Sheffield Ball (born March 29, 1985) is an American football cornerback for the Jacksonville Jaguars of the National Football League. & \multirow{4}{*}{2} \\ \cline{2-2}
$p_{second}^{def}$ & Alan Sheffield Ball (born March 29, 1985) is a former American football cornerback in the National Football League for the Dallas Cowboys, Houston Texans, Jacksonville Jaguars, and Chicago Bears. & \\ \bottomrule
\end{tabular}

%}
\caption{Examples of WikiTiDe for each label. In these specific examples, there is full agreement between all ChatGPT instances that performed the annotation.}
\label{tab:examples}
\end{table*}

Unsuprisingly, thus, there is a significant body of work concerned with, for instance, updating language models by pretraining them on in-domain data \cite{gururangan2020don}, editing specific facts \cite{de2021editing,zhu2020modifying,dai2021knowledge}, continual learning \citep{agarwal2021temporal,del2018short,giulianelli2020analysing,dhingra2022time,loureiro2022timelms}, pre-training with an objective specifically designed to handle infusion of newly coined terms \cite{yu2021dict}, or directly modifying the attention mechanism to account for temporality \cite{rosin2022temporal}. All these, in addition to the extensive body of work on diachronic and dynamic (contextualized and static) word embeddings \cite{hamilton2016cultural,rudolph2016exponential,hamilton2016diachronic,rudolph2018dynamic,hofmann2020dynamic}.

%we find some different attempts to incorporate time as a parameter in their models/tasks such as in works of Hamilton et al.(\cite{hamilton2016cultural},\cite{hamilton2016diachronic})(embeddings trained on different time periods), Hofmann et al. (\cite{hofmann2020dynamic})(temporal and social contexts put in the embeddings), Rudolph and Blei (\cite{rudolph2018dynamic}) (extending  embeddings over a time slice) etc.

Regardless of the method, however, a critical component of time-aware NLP is to have access to dynamically changing facts about language and the world so that LMs are exposed to them. As \citet{jang2022temporalwiki} argues, collaborative resources such as Wikipedia or Wikidata can satisfy this desideratum, since they provide a dynamically updated\footnote{According to \url{https://en.wikipedia.org/wiki/Wikipedia:Statistics}, Wikipedia is edited twice per second. } \textit{life-long} resource. Given this, with \textsc{WikiTiDe} we put forward a benchmark comprised of definition pairs annotated in terms of whether they are the same or not, and if not, if this difference can be attributed to a fundamental change in that term, event or entity (as opposed to, for instance, semantic variations such as introduction of a paraphrase or stylistic nuances). We construct \textsc{WikiTiDe} in a weakly supervised manner via bootstrapping, and evaluate a number of LM-based baselines on the task of determining the type of difference between two timestamped definitions. Our results suggest that bootstrapping is helpful, and that this dataset can be used for both aiding in lexical semantics tasks, as well as for efficient scanning for critical updates in Wikipedia.

%Datasets for evaluation and for pretraining, their connection with temporal NLP. The limitations of these datasets is that they fail to capture critical information that is often only captured in the definition of a term. Because of the rate at which Wikipedia updates its pages, and because of its writing guidelines, we argue that any fundamental update in what we know about something, if it's important enough, will make it into the definition. We build on this to build a dataset which we build fully automatically and evaluate here and there.

%The main contribution of our work can be enumerated by the following.
%\begin{enumerate}
%    \item We introduce a seed dataset containing 3000 labelled pairs of (timestamp,definition) for temporal definition classification task. We also create a unalabelled dataset of 7000 definition pairs for our task.
%    \item We use weak supervision to iteratively train our model, with an aim of creating a robust temporal definition classifier. We show that iteratively bootstrapping a model can achieve good results as the data is increased. We hope that this will motivate to train language models on a smaller seed dataset and gradually improve its performance through a bigger dataset.
%\end{enumerate}

\section{Related Work}
\label{sec:relatedwork}

%Definitions: definition extraction, definition modeling, definitions used for augmenting LMs, definition-based datasets.

This paper can be broadly positioned within two areas, namely lexicograhpic \textit{definitions} (understood as a lexicographic resource but also as a high quality source of information for augmenting LMs), and \textit{diachronic NLP}. We therefore make a clear distinction between them in the review of relevant works.

\paragraph{\noindent \textbf{Definitions}} 
 Definitions have traditionally played a crucial role in NLP and computational lexicography. As the building blocks of dictionaries and encyclopedias, they are used when the meaning of a word is sought \cite{navigli2010learning}, and thus the task of automatically constructing glossaries and terminologies is a well established task in NLP and Information Retrieval \cite{anke2018syntactically,spala2019deft,spala2020semeval,veyseh2020joint,azarbonyad2023generating}. 

However, definitions have also been leveraged to improve the quality of NLP systems. For instance, \citet{bovi2015large} and \citet{espinosa2016extasem} harnessed definitions to build knowledge bases by extracting semantic relations from them; \citet{joshi2020contextualized} used definitions to provide additional context to LMs in reading comprehension tasks; \citet{yu2021dict} pre-trained BERT on tasks that exploit definitions, specifically seeking to improve contextual representations of rare terms; and \citet{xutaxoprompt} used definitions as the backbone of prompt-based taxonomy learning. 

In a parallel strand of work, others have explored \textit{definition modeling} systems (i.e., given a term and potentially some context, generate a definition) \cite{ga:18,zh:19,mi:19,mickus2022semeval,be:20}, and these systems have been applied in tasks such as \textit{controlled} definition modeling, e.g., jargon or varying technical complexity \cite{august2022generating,huang2022understanding}, as well as lexical semantics tasks like word sense disambiguation and word-in-context classification \cite{pilehvar2019wic}.

\paragraph{\noindent \textbf{Diachronic NLP}} While there is agreement in that continual learning helps to mitigate the fundamental issues of temporal misalignment \citep{jang2022temporalwiki} and catastrophic forgetting \citep{cossu2022continual}, the availability of benchmarks for retrieving new facts and evaluating LMs on their capacity to account for them is not overwhelming. Social media seems to be a particularly well suited domain for exploring temporal generalization, given its naturally fast-paced nature, and so we find a number of Twitter-specific benchmarks \cite{osborne2014exponential,yogatama2014dynamic}. Moreover, other resources such as arXiv papers \cite{lazaridou2021mind} or Wikipedia \cite{jang2022temporalwiki} have been benchmarked for evaluating temporal generalization, as well as temporal variations of existing relation extraction datasets \cite{dhingra2022time}.

In this context, we argue that Wikipedia is indeed a valuable and underutilized resource for training and evaluating LMs on their language and knowledge update capabilities. While, as \citet{jang2022temporalwiki} points out, not all changes in Wikipedia or Wikidata correspond to an actual change in the real world, we aim to alleviate this limitation by focusing on changes in definitions alone. In this way, we drastically reduce the chances of falsely confusing one superfluous change in a Wikipedia entry with a change that results in a necessary update of our understanding of a concept or entity. In what follows, we discuss how we create our seed for the \textsc{WikiTiDe} dataset, the algorithm for growing it, and then report on several experimental evaluation results.

\begin{algorithm}[!t]
\caption{Collect Definition Pairs}\label{algorithm}
\begin{algorithmic}[1]
\State Let $P$ be the set of Wikipedia pages
\State Let $D$ be the list of definition pairs
\State Let $n$ be the desired number of definition pairs ($n=10,000$)
\State Let $\textnormal{SRP}(p,tl)$ be a function for \emph{selecting a random page} given a specific timeline $tl$
\State $D=\{\}$

\While{$|D| < n$}
    \State Find a random $p \in P$ with timeline $tl_{y}$
    %\State $t \gets \textnormal{ExtractTimeline}(p)$
    \State $tl_{y} \gets \textnormal{SortYearsAscending}(tl_{y})$
    \State $m \gets \textnormal{FindMedian}(t)$
    \State $p_{first} \gets \textnormal{SRP}(p, tl_{y} \leq m)$
    \State $p_{second} \gets \textnormal{SRP}(p, tl_{y} \geq m)$
    \State $p_{first}^{def} \gets \textnormal{GetDefinition}(p_{first})$
    \State $p_{second}^{def} \gets \textnormal{GetDefinition}(p_{second})$
    \State $D \gets D \cup \{(p_{first}^{def}, p_{second}^{def})\}$
\EndWhile
\end{algorithmic}
\label{alg:getpairs}
\end{algorithm}

\section{\textsc{WikiTiDe}}
\label{sec:dataset}

In this section, we discuss, first, the process of retrieving candidate definition pairs for annotation. Then, we provide details about the annotation process, and finally, present examples and summary statistics, aimed to shed light on the properties of \textsc{WikiTiDe}.

%$\subsection{Extracting candidate pairs}

%Given a set of Wikipedia articles $P$, Wikipedia timeline $T = \{2001, \dots, 2022\}$

The process of creating the required definition pairs of \textsc{WikiTiDe} is shown in Algorithm \ref{alg:getpairs}. In a nutshell, we start from the set $P$ of Wikipedia pages, and construct, by sampling two sufficiently distant definitions (that is, the first sentence of a Wikipedia article $p \in P$), a dataset $D$ which contains 10,000 unannotated definition pairs. After this, we randomly select 30\% from $D$ for annotation, which we perform combining the annotations of 4 instances of GPT-3 \cite{brown2020language}\footnote{Specifically, the version powering ChatGPT: {\small{\texttt{gpt-3.5-turbo}}}.}. The main motivation for ``replacing'' manual annotation with a LM is twofold. First, we posit that we can leverage the knowledge embedded in ChatGPT's parameters about well known entities, concepts and events (well known because they have a corresponding Wikipedia page). Second, recent work has shown that leveraging ChatGPT can outperform other annotation frameworks, for example Amazon Mechanical Turk \cite{gilardi2023chatgpt}. The four rounds of annotations we perform differ in the instruction, as the hyperparameters remain fixed (specifically $temperature=0$ and $top\_{p} = 1$). The instruction combines a prompt and a few examples (potentially - but not always - covering all possible labels). The specific variations involve paraphrasing some of the instructions or definitions of labels, or selecting different examples\footnote{One example of a prompt is provided in the appendix of this submission.}. As for the labels, we define our task as a 3-label classification problem, and hence the 3 different labels (and how they are described to ChatGPT) can be broadly defined as follows:

\begin{enumerate}
    \item Class \textbf{0}: $p_{first}^{def}$ and $p_{second}^{def}$ essentially convey the same information, with negligible differences in terms of style.
    \item Class \textbf{1}: $p_{first}^{def}$ and $p_{second}^{def}$ may be semantically similar but conveying analogous information, or else convey different information, however these differences cannot be attributed to a fundamental change or update in our understanding about $p$.
    \item Class \textbf{2}: $p_{first}^{def}$ and $p_{second}^{def}$ are different, \textit{and} this difference can be unequivocally attributed to some fundamental changes happening to $p$ and/or our shared understanding of $p$, which changed during the period that spanned between $p_{first}^{def}$ and $p_{second}^{def}$.
\end{enumerate}

%hsuvas edit
%{\color{red}
The final labels are selected as follows: We only select instances labeled as class \textbf{2} if all instances of ChatGPT label it as such, thus ensuring the tightest possible agreement for this label, which is both the most interesting and infrequent in the dataset. Then, for the rest, we resort to the label assigned by the majority among three ChatGPT annotators, and only in case of draw, we incorporate a fourth one, which acts as referee. At the end of this process, we annotate 3,000 instances out of the 10,000 initial set, with a Fleiss-Kappa Agreement score of \citep{fleiss1971measuring} of 24.84, which according to the literature, falls within the \textit{fair} agreement range. Table \ref{tab:examples} shows illustrative examples of definition pairs in \textsc{WikiTiDe}. This 3k \textit{training set} ($TS$) has the following label distribution: 1,082 examples for label 0; 1,830 for label 1; and 87 definition pairs for the most interesting label 2. In the following section we describe how we use $TS$ to fully annotate $D$.
%}

%and explained hereforth. We consider Wikipedia as our data source and leverage MediaWiki API to extract the definition pairs used in our experiments. We extract 10,000 articles in a set of timestamps in a yearly manner, that is one version of the article for every year since it's inception, ideally the last revision for that year. After extracting the timestamps, we create the data set by considering two random revisions from the article timeline divided in halves. We then extract the revision text for the timestamps, clean it up and extract the first sentences for the timestamps(since Wikipedia guidelines\footnote{\url{https://www.mediawiki.org/wiki/Documentation/Style_guide}} state that the definitions are in first sentences of the text). We obtain pairs of (timestamp, definition). We then divide them into a set of dataset, containing 3000 articles and a set of 7000 unlabelled articles. We prompt the seed dataset to GPT3.5 with four settings and one human annotation to obtain their labels, which are then considered based on majority annotation. The reason behind using GPT3.5 is to leverage their ability to provide human level annotations \citep{gilardi2023chatgpt}, and we obtain a Fleiss-Kappa Agreement \citep{fleiss1971measuring} of 0.2484, and percentage agreement of 99.8\%.  Finally we break down the seed dataset into train and test sets, using scikit learn's train-test split.

\begin{comment}
\begin{figure}
    \includegraphics[width=0.5\textwidth]{Images/Dataset_diagram.png}
    \caption{Steps of designing the Datasets}
    \centering
    \label{fig:dataset}
\end{figure} 
   
\end{comment}

\section{Bootstrapping \textsc{WikiTiDe}}

With $TS$ being the ChatGPT-annotated seed dataset in \textsc{WikiTiDe} (with a label set $L = \{0, 1, 2\}$), let $DS$ be the remaining unannotated 7,000 instances, and $ D = TS \cup DS $. We seek to iteratively bootstrap a development set with ``high confident'' predictions, starting from a seed classifier trained, in a first iteration, only on $TS$. We argue that this approach, which can be traced back to applications in word sense disambiguation and definition extraction \cite{yarowsky1995unsupervised,Anke2015WeaklySD}, can be effectively applied to our use case as each newly bootstrapped definition pair will be reliable indicatives of the source training set, which can contribute to increase recall as the model will have seen more positive examples.

\begin{algorithm}
\caption{Bootstrapping on \textsc{WikiTiDe}}
\begin{algorithmic}[1]
\Require{Initial training set $TS$}
\Require{Development set $DS$}
\Require{Held-out test set $HS$}
\Require{Label set $L = \{0, 1, 2\}$}
\Require{$\textnormal{K} \gets 10$}

\While{$|DS| \geq \textnormal{topnPreds} \cdot |L|$}
    %\State Train model on $TS$
    \State $\textnormal{model} \gets \textnormal{trainModel}(TS)$    
    %\State Appply model to $DS$
    \State $\textnormal{model}(DS)$ // Apply model to $DS$
    \For{$l \in L$}
        \State $DS_l \gets \{x \mid x \in DS \textnormal{, } \textnormal{label}(x) = l\}$
        \State $P_l \gets \{P(x, l) \mid x \in DS_l\}$
        %\State $P_l' = \frac{{\exp\left(\frac{{\log(P_l)}}{T}\right)}}{{\sum_{i=1}^{|P_l|}\exp\left(\frac{{\log(P_l[i])}}{T}\right)}}$ 
        \State Sort $P_l'$ in descending order
        \State $DS_l' \gets$ Top $\textnormal{K}$ instances from $DS_l$ based on $P_l'$
        \State $TS \gets TS \cup DS_l'$
        \State $DS \gets DS \setminus DS_l$    
    \EndFor
    \State $\textnormal{evaluateModel}(\textnormal{model},HS)$
\EndWhile
\end{algorithmic}
\label{alg:bootstrapping}
\end{algorithm}

As summarized in Algorithm \ref{alg:bootstrapping}, the bootstrapping process requires at a minimum an annotated training set $TS$ and an unannotated test set $DS$, and optionally held-out test set $HS$ to monitor performance. At the first iteration, we set $|TS|=2160; |DS|=7,000;$ and $|HS|=840$. We then fire the bootstrapping process, in which, first, a model is trained and applied on $DS$, then we extract the $K$ most confident predictions for each label, append them to $TS$, and remove them from $DS$. Every time we exhaust all labels in $L$, we evaluate a new instance of the model on $HS$. 
%Finally, in order to reliably use model probabilities as a proxy for confidence, we apply temperature scaling \cite{pmlr70guo17a} (line 7 in Algorithm \ref{alg:bootstrapping}). 

In terms of classifier, we select a wide range of models to evaluate, all of them based on the Transformers architecture \cite{vaswani2017attention}, namely BERT \cite{devlin-etal-2019-bert}, RoBERTa \cite{liu2019roberta}, DistilBERT and DistilroBERTa \citep{sanh2020distilbert}, Tiny-BERT \citep{bhargava2021generalization,DBLP:journals/corr/abs-1908-08962} and XLM-Roberta-base \citep{DBLP:journals/corr/abs-1911-02116}\footnote{All of them available at the Huggingface model hub \url{www.huggingface.co}.}. Finally, in terms of manipulating the inputs to these models, we opt for minimal preprocessing, simply injecting special tokens `$<$ y$>$' and `$<$/y$>$' for isolating timespans, and `$<$t$>$' and `$<$/t$>$' in order to mark the target term. % We use the Huggingface implementation of the models and keep the same training configuration for all of them, to maintain fair comparisons.

\begin{table*}[!t]
\resizebox{\textwidth}{!}{% 
\begin{tabular}{@{}ll|rrr|rrr|rrr|rrr|r@{}}
\toprule 
                      &              & \multicolumn{3}{c}{\texttt{Label 2}}                                                  & \multicolumn{3}{|c}{\texttt{Label 1}}                                                  & \multicolumn{3}{|c}{\texttt{Label 0}}                                                  & \multicolumn{3}{|c|}{\texttt{Avg.}}                                                        & \multicolumn{1}{l}{}                                                       \\ \midrule
Model                 & \textbf{Boot.} & \multicolumn{1}{|c}{\textbf{P}} & \multicolumn{1}{c}{\textbf{R}} & \multicolumn{1}{c}{\textbf{F1}} & \multicolumn{1}{|c}{\textbf{P}} & \multicolumn{1}{c}{\textbf{R}} & \multicolumn{1}{c}{\textbf{F1}} & \multicolumn{1}{|c}{\textbf{P}} & \multicolumn{1}{c}{\textbf{R}} & \multicolumn{1}{c}{\textbf{F1}} & \multicolumn{1}{|c}{\textbf{\textbf{P}}} & \multicolumn{1}{c}{\textbf{R}} & \multicolumn{1}{c}{\textbf{F1}} & \multicolumn{1}{|l}{\textbf{BI}} \\ \midrule
roberta-base          & no            &48.98     &50.00&49.49          & 77.47&77.67&77.56         &78.94&79.69&79.24          &68.47&69.12&68.76                                    \\ 
roberta-base          & yes           &81.22&70.35&74.58         &86.93&\textbf{88.33}&87.08         &88.07&\textbf{90.13}&88.43          &85.41&\textbf{82.94}&83.62         &47                                 \\\midrule
distilbert-base-cased & no            &64.83&72.44&67.78         &75.96&76.98&75.81         &77.79&79.33&78.05          &72.88&76.25&73.88                                    \\ 
distilbert-base-cased & yes           &74.28&64.40&68.00         &80.16&81.34&80.12         &81.44&83.22&81.54          &78.62&76.32&76.56         &28                                 \\\midrule
xlm-roberta-base      & no            &48.99&50.00&49.49         &29.88&50.00&37.41         &30.89&50.00&38.19          &36.59&50.00&41.70                                    \\ 
xlm-roberta-base      & yes           &67.91&58.52&61.43         &84.72&86.09&84.61         &86.53&88.65&86.56          &79.72&77.75&77.53         &9                                  \\\midrule
bert-base-cased       & no            &60.97&60.97&60.97         &59.84&53.57&48.33         &65.68&54.67&49.62          &62.17&56.41&52.98                                    \\ 
bert-base-cased       & yes            &63.73&\textbf{72.31}&66.89         &72.24&73.12&72.07         &73.60&74.83&73.74          &69.86&73.42&70.90&14                                    \\ \midrule
bert-tiny              & no            &48.76&40.77&44.41         &51.09&50.86&49.73         &41.36&47.46&39.91          &47.07&46.36&44.68                                    \\
bert-tiny              & yes           &50.80&52.54&50.52         &57.42&57.15&57.19         &57.66&56.68&56.72          &55.29&55.49&54.81         &44                                 \\\midrule
distilroberta-base     & no            &48.99&50.00&49.49         &73.38&73.88&71.64         &75.23&76.24&73.14          &65.87&66.71&64.76                                    \\
distilroberta-base     & yes           &60.86&66.43&63.01         &80.67&81.88&80.52         &83.05&84.84&83.33          &74.86&77.72&75.61         &11                                 \\\midrule
roberta-large          & no            &48.99&50.00&49.49         &81.03&64.34&62.86         &82.19&65.15&64.57          &70.74&57.17&58.97                                    \\
roberta-large          & yes           &\textbf{88.29}&70.47&\textbf{76.56}         &\textbf{87.59}&88.25&\textbf{87.86}         &\textbf{88.76}&89.90&\textbf{89.21}          &\textbf{88.21}&82.87&\textbf{84.54}         &54                                 \\ \bottomrule 
\end{tabular}%
}
\caption{Results on the held-out test set $HS$ for a number of LMs. For the bootstrapped models, we also report the best iteration (column \textbf{BI}).} 
\label{tab:main-results}
\end{table*}

\subsection{Results and Discussion}

We flesh out the results obtained by different models in the task of predicting, given a pair of definitions from Wikipedia, the labels introduced in Section \ref{sec:dataset}. As can be seen in Table \ref{tab:main-results}, the bootstrapped models are consistently better than their base counterparts (which, we recall, are equivalent models but being trained only on $TS$). RoBERTa-based models are superior to the rest, and crucically, they also reach to the best performing iteration at later stages, which suggests they tend to overfit less to the training set. In terms of gap between base and boostrapped models, this is rather large, and largest for label \textbf{2}. As an example, RoBERTa-large is almost 40 points more precise when bootstrapped, and 27 F1 points better. Interestingly, our intuition of using a multilingual model to handle ``foreign'' (non English) spellings, typically used in Wikipedia definitions for non English entities or concepts, seems to not work well, with XLM-roBERTa-base being the 2nd to last model, only surpassing BERT-tiny. 

\begin{figure}
    \includegraphics[width=0.5\textwidth]{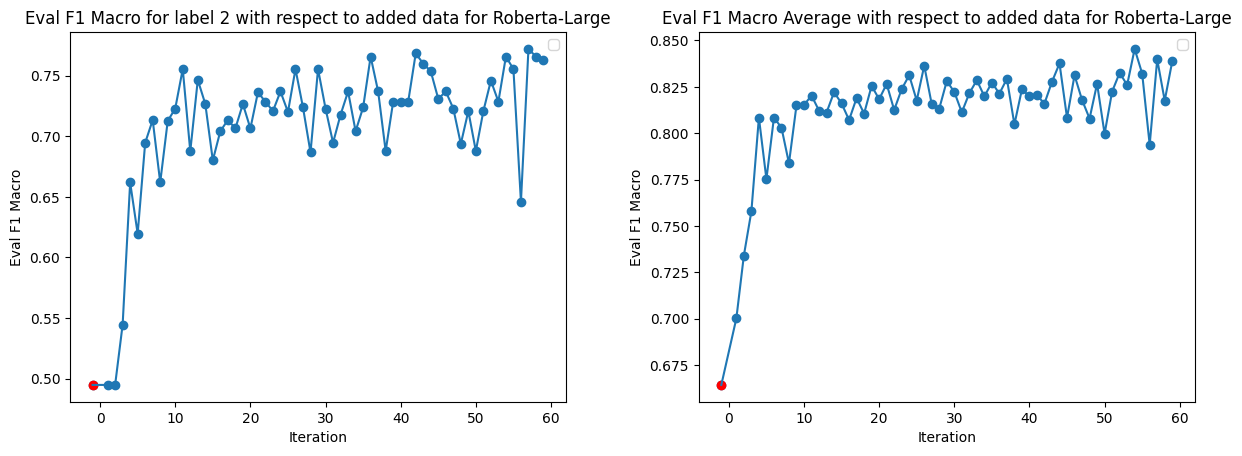}
    \caption{Macro-F1 scores of Roberta-Large with respect to Number of Iterations}
    \centering
    \label{fig:roberta_acc}
\end{figure}

\begin{figure}
    \includegraphics[width=0.5\textwidth]{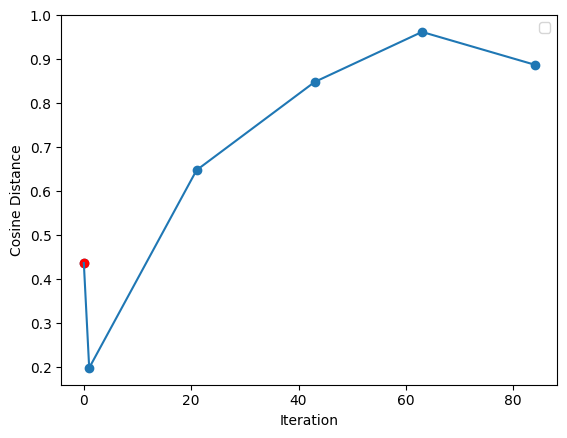}
    \caption{Cosine Distance of definition pairs for Label 2 with respect to bootstrapping iterations}
    \centering
    \label{fig:cosine-label2}
\end{figure}

In terms of analyzing the bootstrapping iterative process, we can see in Figure \ref{fig:roberta_acc} that the improvements of the bootstrapped models becomes apparent after few iterations, both for the most relevant label 2 (left plot) and on average (right plot). We also see less ``up and down spikes'' for the average results, suggesting that performance on the other labels becomes smoother over time. 
Moreover, in order to gain further understanding on the effects of the bootrsapping process into the differences in definition pairs over time, we measure \textit{semantic drift}, i.e., whether (or, more precisely, the extent to which) the bootstrapped training set exhibits an increasingly diverse set of definitions, measured by how dissimilar they are as they are iteratively fetched from $DS$. We focus on label 2, and plot the results of this analysis in Figure \ref{fig:cosine-label2}, which clearly shows an increasing drift in average distances. This confirms that the bootstrapped training set is semantically more diverse than the seed ChatGPT-annotated version.

\begin{table*}[!t]
\centering
%\resizebox{\textwidth}{!}{%
\small
\begin{tabular}{|p{0.3cm}|p{1cm}|p{11cm}|m{0.5cm}|}
\toprule
\multicolumn{1}{c}{Iteration} & \multicolumn{1}{c}{} & \multicolumn{1}{|c|}{WikiTiDe Definitions} & \multicolumn{1}{r}{Label} \\ \midrule
\multirow{2}{*}{1} & $p_{first}^{def}$ & Argentine football saw Lomas Athletic Club win their 5th Argentine championship in 6 seasons & \multirow{2}{*}{2} \\ \cline{2-2}
 & $p_{second}^{def}$ &  Argentine football saw Lomas win its 5th Primera División championship within 6 seasons. & \\ \midrule
\multirow{2}{*}{1} & $p_{first}^{def}$ & The 7th Army Aviation Regiment is an army aviation formation of the Ukrainian Ground Forces & \multirow{2}{*}{1} \\ \cline{2-2}
& $p_{second}^{def}$ & The Army Aviation Brigade is an army aviation formation of the Ukrainian Ground Forces. & \\ \midrule
 \multirow{2}{*}{1} & $p_{first}^{def}$ & The 16S rRNA is a long component of the small prokaryotic ribosomal subunit (30S) and is known to interact with the 50S subunit in both P and A site. & \multirow{2}{*}{0} \\ \cline{2-2}
& $p_{second}^{def}$ & 16S ribosomal RNA (or 16S rRNA) is the RNA component of the 30S subunit of a prokaryotic ribosome (SSU rRNA). & \\ \midrule \midrule
\multirow{2}{*}{43} & $p_{first}^{def}$ & Dr. Bhupendranath Dutta was a famous Indian revolutionary and later a noted Sociologist. & \multirow{2}{*}{2} \\ \cline{2-2}
 & $p_{second}^{def}$ &  Bhupendranath Datta was an Indian revolutionary and later a noted sociologist and anthropologist. & \\ \midrule
\multirow{2}{*}{43} & $p_{first}^{def}$ & Dexia Mons-Hainaut is the Belgian professional basketball club, who based in Quaregnon. & \multirow{2}{*}{1} \\ \cline{2-2}
& $p_{second}^{def}$ & Belfius Mons-Hainaut is a Belgian professional basketball club that is based in Mons, Wallonia. & \\ \midrule
 \multirow{2}{*}{43} & $p_{first}^{def}$ & Berkshire soil series is the name given to a well drained loam or sandy loam soil which has developed on glacial till in parts of southern Quebec, eastern New York State and New England south to Massachusetts. & \multirow{2}{*}{0} \\ \cline{2-2}
& $p_{second}^{def}$ & Berkshire soil series is the name given to a well-drained loam or sandy loam soil which has developed on glacial till in parts of southern Quebec, eastern New York State and New England south to Massachusetts. & \\ 
\midrule \midrule
\multirow{2}{*}{83} & $p_{first}^{def}$ &  Carlos Alberto Valencia is a Colombian left wing back who plays for River Plate of Buenos Aires, Argentina. & \multirow{2}{*}{2} \\ \cline{2-2}
 & $p_{second}^{def}$ &   Carlos Alberto Valencia Paredes is a Colombian footballer who plays as a left-back for Independiente Medellín. & \\ \midrule 
\multirow{2}{*}{83} & $p_{first}^{def}$ & The Carnegie Free Library of Beaver Falls  was the first public library built in Beaver County, Pennsylvania.  & \multirow{2}{*}{1} \\ \cline{2-2}
& $p_{second}^{def}$ & The Carnegie Free Library of Beaver Falls  is a historic Carnegie library in the city of Beaver Falls, Pennsylvania, United States. & \\ \midrule

 \multirow{2}{*}{83} & $p_{first}^{def}$ & Carl-Johan Lindqvist is a Swedish luger who competed in the early 1990s & \multirow{2}{*}{0} \\ \cline{2-2}
& $p_{second}^{def}$ & Carl-Johan Alexander Lindqvist (born November 15, in Tyresö) is a Swedish luger who competed in the early 1990s. & \\\bottomrule

\end{tabular}
%}
\caption{Examples of Model output on different iterations of Bootstrapping for Roberta-Base.}
\label{tab:model-iter-res}
\end{table*}

As a form of qualitative evaluation, we list in Table \ref{tab:model-iter-res} a set of bootstrapped instances from one of the best performing models (RoBERTa-base). Note that these are not carefully selected examples, as we have simply listed an instance of high confidence classifications per label. We can see the improvement in quality of 2-labeled instances, especially between iterations 1 and 83, in which the difference in knowledge concerning Carlos Alberto Valencia is minimal in terms of string edit distance, however the model correctly identified a critical change for this named entity, specifically, the fact that he changed teams.

\begin{table*}[!t]
\small
\resizebox{\textwidth}{!}{%
\begin{tabular}{lrrrrrrrrrrrr}
\toprule
                      & \multicolumn{4}{c}{Train}                                                                                                                       & \multicolumn{4}{c}{Dev} 
                                & \multicolumn{4}{c}{Test} 
                      \\ \midrule
                      & \multicolumn{2}{c}{Original}                                            & \multicolumn{2}{c}{GPT3.5}                                             & \multicolumn{2}{c}{Original}                                            & \multicolumn{2}{c}{GPT3.5}    
                                    & \multicolumn{2}{c}{Original}                               & \multicolumn{2}{c}{GPT3.5}      
                                            \\
                      & \multicolumn{1}{c}{Base} & \multicolumn{1}{c}{Bootsr.} & \multicolumn{1}{c}{Base} & \multicolumn{1}{c}{Bootsr.} & \multicolumn{1}{c}{Base} & \multicolumn{1}{c}{Bootsr.} & \multicolumn{1}{c}{Base} & \multicolumn{1}{c}{Bootsr.} &
                      \multicolumn{1}{c}{Base} & \multicolumn{1}{c}{Bootsr.} &
                      \multicolumn{1}{c}{Base} & \multicolumn{1}{c}{Bootsr.}\\\midrule
roberta-base          & 0.33                                & 0.48                             & 0.33                            & 0.35                               & 0.34                                & 0.44                          & 0.33                 & 0.34 
    & 0.34                                & 0.50                          & 0.34                                                  & 0.35  \\

distilbert-base-cased & 0.33                                & 0.46                             & 0.33                            & 0.33                                & 0.3                                & 0.47                             & 0.34                              & 0.34                                & 0.34                                & 0.43                         & 0.34                 & 0.34     \\

xlm-roberta-base      & 0.33                                & 0.39                             & 0.33                            & 0.40                                & 0.34                                & 0.34                        & 0.34                             & 0.34                                & 0.34                                & 0.34                         & 0.34                 & 0.43      \\

bert-base-cased       & 0.3                                & 0.38                             & 0.33                                  & 0.39                                & 0.46                                & 0.52                             & 0.3                              & 0.34                               & 0.48                                & 0.48                        & 0.34                 & 0.43      \\

bert-tiny             & 0.33                                & 0.33                             & 0.33                                   & 0.35                                & 0.3                                & 0.34                             & 0.33                              & 0.34                        & 0.36                                & 0.36                         & 0.37                 & 0.35       \\

distilroberta-base    & 0.34                                & 0.34                            & 0.33                                   & 0.34                                 & 0.34                                & 0.34                             & 0.33                             & 0.34                       & 0.36                                & 0.36                         & 0.34                 & 0.35         \\

roberta-large         & 0.30                                & 0.53                             & 0.33                                   & 0.50                                & 0.34                                & 0.51                             & 0.34                             & 0.48                              & 0.34                                & 0.45                        & 0.34                 & 0.45  \\\midrule\midrule
                      & \multicolumn{4}{c}{Train}                                                                                                                       & \multicolumn{4}{c}{Dev}                                                  & \multicolumn{4}{c}{Test}                                                                 \\\midrule
                      & \multicolumn{2}{c}{Original}                                            & \multicolumn{2}{c}{GPT3.5}                                             & \multicolumn{2}{c}{Original}                                            & \multicolumn{2}{c}{GPT3.5} 
                                    & \multicolumn{2}{c}{Original}                
                                      & \multicolumn{2}{c}{GPT3.5} \\
                      & \multicolumn{1}{c}{Base} & \multicolumn{1}{c}{Bootsr.} & \multicolumn{1}{c}{Base} & \multicolumn{1}{c}{Bootsr.} & \multicolumn{1}{c}{Base} & \multicolumn{1}{c}{Bootsr.} & \multicolumn{1}{c}{Base} & \multicolumn{1}{c}{Bootsr.} &
                      \multicolumn{1}{c}{Base} & \multicolumn{1}{c}{Bootsr.} &
                      \multicolumn{1}{c}{Base} & \multicolumn{1}{c}{Bootsr.}\\\midrule
roberta-base          & 0.50                                & 0.50                             & 0.50                                   & 0.50                                & 0.51                                & 0.48                             & 0.50                                   & 0.51                             & 0.51                                & 0.53                         & 0.51                 & 0.51   \\

distilbert-base-cased & 0.5                                & 0.49                             & 0.49                                   & 0.50                                & 0.51                                & 0.50                            & 0.51                              & 0.50                             & 0.51                                & 0.48                         & 0.50                 & 0.51   \\

xlm-roberta-base      & 0.50                                & 0.50                             & 0.50                                & 0.49                                   & 0.5                                & 0.51                           & 0.50                                & 0.51                           & 0.50                               & 0.51                         & 0.51                 & 0.50     \\

bert-base-cased       & 0.50                                & 0.52                             & 0.50                                   & 0.52                                & 0.50                                & 0.50                             & 0.51                                  & 0.51                         & 0.49                                & 0.49                         & 0.51                 & 0.51       \\

bert-tiny             & 0.50                                & 0.50                             & 0.50                                   & 0.50                                & 0.51                               & 0.51                             & 0.40                                   & 0.5                          & 0.49                                & 0.50                         & 0.48                 & 0.50     \\

distilroberta-base    & 0.50                                & 0.50                             & 0.50                                   & 0.50                                & 0.50                                & 0.50                             & 0.51                             & 0.52                              & 0.49                                & 0.50                         & 0.51                 & 0.51  \\

roberta-large         & 0.50                                & 0.51                             & 0.49                                   & 0.51                                & 0.50                                & 0.51                             & 0.50                                   & 0.49  & 0.50                                & 0.46                         & 0.51                 & 0.48      \\ \bottomrule                             
\end{tabular}%
}
\caption{F1 (top) and accuracy (bottom) results on WiC-TSV. The Vanilla columns refer to instances where we run inference with a classifier trained on \textsc{WikiTiDe} directly, without adapting inputs or further fine-tuning. GPT3.5 columns denote a use case where we use GPT3.5 for generating a definition of the target word in the first sentence of the dataset instance, and then run inference on this updated input.}
\label{tab:wictsv}
\end{table*}
\section{Case Study: WiC-TSV}

The WiC-TSV (Word in Context-Target Sense Verification) task \cite{breit2021wic} is a ``shootoff'' from the original WiC task \cite{pilehvar2019wic}. It proposes a binary classification problem, where the input is a pair of sentences: the first one, a sentence with a target word in context, and the second one, a definition of that target word. This is a suitable test bet for a model fine-tuned on \textsc{WikiTiDe}, since this is a dataset which essentially measures definition similarity. However, since \textsc{WikiTiDe} is a multilabel dataset, we combine labels 1 and 2 as label 0 in WiC-TSV and assume equivalence between the notion of ``change'' in \textsc{WikiTiDe} and polysemy in WiC-TSV. For our model to work, both input sentences must be definitions, however, this is not always the case in WiC-TSV. To work around this limitation, we replace the non-definition sentences in WiC-TSV with a definition generated using ChatGPT \cite{brown2020language}. Both sets of results (directly applying our model to WiC-TSV as well as replacing one of its sentences with a ChatGPT-generated definition) are reported, for train, test and development sets\footnote{\url{https://github.com/semantic-web-company/wic-tsv/tree/master/data/en}.} (which is possible as we cast this problem as an unsupervised classification task), in Table \ref{tab:wictsv}.

Moreover, we report results reported in previous works to further contextualize the results we obtain, which, to reiterate, are from an unsupervised model not directly optimized for this task. \citet{breit2021wic} reports the \textit{all true} baseline on the test split has having Accuracy of 50.8\% and F1 of 67.3\%. Additionally, they obtain Accuracy scores of of 54.4\% and F1 scores of 26.2\% with an unsupervised BERT-based model, whereas they find significant improvements (Accuracy, 76.0\% and F1-score, 78.8\%) for a supervised GBERT-based model. %We report these results from Task1 of their original work, which is definition information task, and relates to our task as well. 
We also find in the work by \citet{zervakis}, where they propose target sense verification as an analogy detection task, that they achieve Accuracy scores of 78.6\% and F1 of 79.7\% on the test set (for supervised approaches), and  Accuracy of 61.2\% (and 51.3\% F1) for an unsupervised approach.

The results of our experiment display the ability of the models before and after bootstrapping on all three sets (train,deveopment and test). The bootstrapped approach considerably increases the Macro-F1 performance of the models with respect to WiC-TSV's Task 1 unsupervised setting baselines \citep{breit2021wic}. The results also suggest that while BERT shines on a few occasions, the RoBERTa family of models show the highest performance, with RoBERTa-large bootstrapped being the best with F1 score of 0.53. The vanilla versions of the models perform within a range between 0.30 to 0.34. The bootstrapped versions outperform their non-bootstrapped counterparts in all three datasets, with respect to F1 score. We also observe that the difference in F1 scores between before and after bootstrapped versions can go as high as 17 points, which signify that the models learns better during the bootstrapping process. %However, in case of accuracy, both versions of model perform similar, with a small performance gap. 
Finally, we also find that the larger models with more parameters outperform their distilled counterparts in most of the dataset versions. 
\begin{comment}
\textcolor{red}
{An example of the responses of the models(bootstrapped/non-b on different test data is given below. \\
Term: Offing\\
Def1:  Offing is the part of the sea that can be seen from the shore and is beyond the anchoring area.\\
Def2:  Offing, a time or event that is imminent or impending.\\
No bootstrapped model Response: False (No change/not different)\\
Bootstrapped Response: True (change/different)\\
Gold label: True (Change/different)
}    
\end{comment}

%The dataset contains a  word, a pair of definitions, and context sentences, accompanied by a label on whether the definition sentence correctly disambiguates the usage of the word in the context sentence. For our experimental purpose, we consider the context sentence as another definition sentence. Notable that this assumption may not hold in many cases, however, it can also be a test of robustness for our classifier. We then run our best-performing models (based on the evaluation mentioned in the previous section) and report the results of our experiment in Table \ref{tab:wic_res}. The table shows that the models in bootstrapped settings perform better than non bootstrapped counterparts  by a significant margin in many cases.

% \usepackage{booktabs}
% \usepackage{graphicx}

\section{Conclusion}
We propose a dataset and methodology to design a classifier for detecting temporal changes in temporal definition pairs. We use weak supervision technique by boostrapping the model an unlabelled dataset in output controlled setting. We also see that bootstrapping a model improves the accuracy of the model as well as makes the model more robust. However, the process requires more time to bootstrap the model and the success of the process depends on the initial training. Although the process has its own limitations, we conclude that the idea of using a classifier to detect information changes in with respect to temporality and training it with boostrapping can result in easement of defining which information is relevant to update a model's knowledge base and can help to mitigate the issues that a language model suffers due to temporal misalignment.

\section*{Ethics and Broader Statement}

This paper is concerned with the automatic construction of a dataset by combining publicly available information in the web. Therefore, it might be possible that incorrect or harmful information is present in this derived dataset, although we welcome efforts by the community to contribute mitigating these risks. The dataset construction process did not involve humans.

Potential risks in the dataset might also include incorrectly flagging new knowledge about any article, as our data source Wikipedia is a publicly editable data source. Therefore the possibility of having conflicting or incorrect information also increases. However, the difference of information, which our classifier is trained to detect can help to detect such outliers and provide some insights about it.

\bibliographystyle{acl_natbib}
\bibliography{anthology,ranlp2023}

\begin{thebibliography}{56}
\expandafter\ifx\csname natexlab\endcsname\relax\def\natexlab#1{#1}\fi

\bibitem[{Agarwal and Nenkova(2021)}]{agarwal2021temporal}
Oshin Agarwal and Ani Nenkova. 2021.
\newblock Temporal effects on pre-trained models for language processing tasks.
\newblock \emph{arXiv preprint arXiv:2111.12790}.

\bibitem[{August et~al.(2022)August, Reinecke, and
  Smith}]{august2022generating}
Tal August, Katharina Reinecke, and Noah~A Smith. 2022.
\newblock Generating scientific definitions with controllable complexity.
\newblock In \emph{Proceedings of the 60th Annual Meeting of the Association
  for Computational Linguistics (Volume 1: Long Papers)}, pages 8298--8317.

\bibitem[{Azarbonyad et~al.(2023)Azarbonyad, Afzal, and
  Tsatsaronis}]{azarbonyad2023generating}
Hosein Azarbonyad, Zubair Afzal, and George Tsatsaronis. 2023.
\newblock Generating topic pages for scientific concepts using scientific
  publications.
\newblock In \emph{Advances in Information Retrieval: 45th European Conference
  on Information Retrieval, ECIR 2023, Dublin, Ireland, April 2--6, 2023,
  Proceedings, Part II}, pages 341--349. Springer.

\bibitem[{Bevilacqua et~al.(2020)Bevilacqua, Maru, and Navigli}]{be:20}
Michele Bevilacqua, Marco Maru, and Roberto Navigli. 2020.
\newblock \href {https://doi.org/10.18653/v1/2020.emnlp-main.585}
  {Generationary or {``}how we went beyond word sense inventories and learned
  to gloss{''}}.
\newblock In \emph{Proceedings of the 2020 Conference on Empirical Methods in
  Natural Language Processing (EMNLP)}, pages 7207--7221, Online. Association
  for Computational Linguistics.

\bibitem[{Bhargava et~al.(2021)Bhargava, Drozd, and
  Rogers}]{bhargava2021generalization}
Prajjwal Bhargava, Aleksandr Drozd, and Anna Rogers. 2021.
\newblock \href {http://arxiv.org/abs/2110.01518} {Generalization in nli: Ways
  (not) to go beyond simple heuristics}.

\bibitem[{Breit et~al.(2021)Breit, Revenko, Rezaee, Pilehvar, and
  Camacho-Collados}]{breit2021wic}
Anna Breit, Artem Revenko, Kiamehr Rezaee, Mohammad~Taher Pilehvar, and Jose
  Camacho-Collados. 2021.
\newblock \href {https://www.aclweb.org/anthology/2021.eacl-main.140}
  {{WiC-TSV}: {A}n evaluation benchmark for target sense verification of words
  in context}.
\newblock In \emph{Proceedings of the 16th Conference of the European Chapter
  of the Association for Computational Linguistics: Main Volume}, pages
  1635--1645, Online. Association for Computational Linguistics.

\bibitem[{Brown et~al.(2020)Brown, Mann, Ryder, Subbiah, Kaplan, Dhariwal,
  Neelakantan, Shyam, Sastry, Askell et~al.}]{brown2020language}
Tom Brown, Benjamin Mann, Nick Ryder, Melanie Subbiah, Jared~D Kaplan, Prafulla
  Dhariwal, Arvind Neelakantan, Pranav Shyam, Girish Sastry, Amanda Askell,
  et~al. 2020.
\newblock Language models are few-shot learners.
\newblock \emph{Advances in neural information processing systems},
  33:1877--1901.

\bibitem[{Chowdhery et~al.(2022)Chowdhery, Narang, Devlin, Bosma, Mishra,
  Roberts, Barham, Chung, Sutton, Gehrmann et~al.}]{chowdhery2022palm}
Aakanksha Chowdhery, Sharan Narang, Jacob Devlin, Maarten Bosma, Gaurav Mishra,
  Adam Roberts, Paul Barham, Hyung~Won Chung, Charles Sutton, Sebastian
  Gehrmann, et~al. 2022.
\newblock Palm: Scaling language modeling with pathways.
\newblock \emph{arXiv preprint arXiv:2204.02311}.

\bibitem[{Conneau et~al.(2019)Conneau, Khandelwal, Goyal, Chaudhary, Wenzek,
  Guzm{\'{a}}n, Grave, Ott, Zettlemoyer, and
  Stoyanov}]{DBLP:journals/corr/abs-1911-02116}
Alexis Conneau, Kartikay Khandelwal, Naman Goyal, Vishrav Chaudhary, Guillaume
  Wenzek, Francisco Guzm{\'{a}}n, Edouard Grave, Myle Ott, Luke Zettlemoyer,
  and Veselin Stoyanov. 2019.
\newblock \href {http://arxiv.org/abs/1911.02116} {Unsupervised cross-lingual
  representation learning at scale}.
\newblock \emph{CoRR}, abs/1911.02116.

\bibitem[{Cossu et~al.(2022)Cossu, Tuytelaars, Carta, Passaro, Lomonaco, and
  Bacciu}]{cossu2022continual}
Andrea Cossu, Tinne Tuytelaars, Antonio Carta, Lucia Passaro, Vincenzo
  Lomonaco, and Davide Bacciu. 2022.
\newblock \href {http://arxiv.org/abs/2205.09357} {Continual pre-training
  mitigates forgetting in language and vision}.

\bibitem[{Dai et~al.(2021)Dai, Dong, Hao, Sui, Chang, and
  Wei}]{dai2021knowledge}
Damai Dai, Li~Dong, Yaru Hao, Zhifang Sui, Baobao Chang, and Furu Wei. 2021.
\newblock Knowledge neurons in pretrained transformers.
\newblock \emph{arXiv preprint arXiv:2104.08696}.

\bibitem[{De~Cao et~al.(2021)De~Cao, Aziz, and Titov}]{de2021editing}
Nicola De~Cao, Wilker Aziz, and Ivan Titov. 2021.
\newblock Editing factual knowledge in language models.
\newblock \emph{arXiv preprint arXiv:2104.08164}.

\bibitem[{Del~Tredici et~al.(2018)Del~Tredici, Fern{\'a}ndez, and
  Boleda}]{del2018short}
Marco Del~Tredici, Raquel Fern{\'a}ndez, and Gemma Boleda. 2018.
\newblock Short-term meaning shift: A distributional exploration.
\newblock \emph{arXiv preprint arXiv:1809.03169}.

\bibitem[{Delli~Bovi et~al.(2015)Delli~Bovi, Telesca, and
  Navigli}]{bovi2015large}
Claudio Delli~Bovi, Luca Telesca, and Roberto Navigli. 2015.
\newblock Large-scale information extraction from textual definitions through
  deep syntactic and semantic analysis.
\newblock \emph{Transactions of the Association for Computational Linguistics},
  3:529--543.

\bibitem[{Devlin et~al.(2019)Devlin, Chang, Lee, and
  Toutanova}]{devlin-etal-2019-bert}
Jacob Devlin, Ming-Wei Chang, Kenton Lee, and Kristina Toutanova. 2019.
\newblock \href {https://doi.org/10.18653/v1/N19-1423} {{BERT}: Pre-training of
  deep bidirectional transformers for language understanding}.
\newblock In \emph{Proceedings of the 2019 Conference of the North {A}merican
  Chapter of the Association for Computational Linguistics: Human Language
  Technologies, Volume 1 (Long and Short Papers)}, pages 4171--4186,
  Minneapolis, Minnesota. Association for Computational Linguistics.

\bibitem[{Dhingra et~al.(2022)Dhingra, Cole, Eisenschlos, Gillick, Eisenstein,
  and Cohen}]{dhingra2022time}
Bhuwan Dhingra, Jeremy~R Cole, Julian~Martin Eisenschlos, Daniel Gillick, Jacob
  Eisenstein, and William~W Cohen. 2022.
\newblock Time-aware language models as temporal knowledge bases.
\newblock \emph{Transactions of the Association for Computational Linguistics},
  10:257--273.

\bibitem[{Espinosa-Anke et~al.(2015)Espinosa-Anke, Saggion, and
  Ronzano}]{Anke2015WeaklySD}
Luis Espinosa-Anke, Horacio Saggion, and Francesco Ronzano. 2015.
\newblock Weakly supervised definition extraction.
\newblock In \emph{Recent Advances in Natural Language Processing}.

\bibitem[{Espinosa-Anke et~al.(2016)Espinosa-Anke, Saggion, Ronzano, and
  Navigli}]{espinosa2016extasem}
Luis Espinosa-Anke, Horacio Saggion, Francesco Ronzano, and Roberto Navigli.
  2016.
\newblock Extasem! extending, taxonomizing and semantifying domain
  terminologies.
\newblock In \emph{Proceedings of the AAAI Conference on Artificial
  Intelligence}, volume~30.

\bibitem[{Espinosa-Anke and Schockaert(2018)}]{anke2018syntactically}
Luis Espinosa-Anke and Steven Schockaert. 2018.
\newblock Syntactically aware neural architectures for definition extraction.
\newblock In \emph{Proceedings of the 2018 Conference of the North American
  Chapter of the Association for Computational Linguistics: Human Language
  Technologies, Volume 2 (Short Papers)}, pages 378--385.

\bibitem[{Fleiss(1971)}]{fleiss1971measuring}
Joseph~L Fleiss. 1971.
\newblock Measuring nominal scale agreement among many raters.
\newblock \emph{Psychological bulletin}, 76(5):378.

\bibitem[{Gadetsky et~al.(2018)Gadetsky, Yakubovskiy, and Vetrov}]{ga:18}
Artyom Gadetsky, Ilya Yakubovskiy, and Dmitry Vetrov. 2018.
\newblock \href {https://doi.org/10.18653/v1/P18-2043} {Conditional generators
  of words definitions}.
\newblock In \emph{Proceedings of the 56th Annual Meeting of the Association
  for Computational Linguistics (Volume 2: Short Papers)}, pages 266--271,
  Melbourne, Australia. Association for Computational Linguistics.

\bibitem[{Gilardi et~al.(2023)Gilardi, Alizadeh, and
  Kubli}]{gilardi2023chatgpt}
Fabrizio Gilardi, Meysam Alizadeh, and Maël Kubli. 2023.
\newblock \href {http://arxiv.org/abs/2303.15056} {Chatgpt outperforms
  crowd-workers for text-annotation tasks}.

\bibitem[{Giulianelli et~al.(2020)Giulianelli, Del~Tredici, and
  Fern{\'a}ndez}]{giulianelli2020analysing}
Mario Giulianelli, Marco Del~Tredici, and Raquel Fern{\'a}ndez. 2020.
\newblock Analysing lexical semantic change with contextualised word
  representations.
\newblock \emph{arXiv preprint arXiv:2004.14118}.

\bibitem[{Gururangan et~al.(2020)Gururangan, Marasovi{\'c}, Swayamdipta, Lo,
  Beltagy, Downey, and Smith}]{gururangan2020don}
Suchin Gururangan, Ana Marasovi{\'c}, Swabha Swayamdipta, Kyle Lo, Iz~Beltagy,
  Doug Downey, and Noah~A Smith. 2020.
\newblock Don't stop pretraining: Adapt language models to domains and tasks.
\newblock \emph{arXiv preprint arXiv:2004.10964}.

\bibitem[{Hamilton et~al.(2016{\natexlab{a}})Hamilton, Leskovec, and
  Jurafsky}]{hamilton2016cultural}
William~L. Hamilton, Jure Leskovec, and Dan Jurafsky. 2016{\natexlab{a}}.
\newblock \href {https://doi.org/10.18653/v1/D16-1229} {Cultural shift or
  linguistic drift? comparing two computational measures of semantic change}.
\newblock In \emph{Proceedings of the 2016 Conference on Empirical Methods in
  Natural Language Processing}, pages 2116--2121, Austin, Texas. Association
  for Computational Linguistics.

\bibitem[{Hamilton et~al.(2016{\natexlab{b}})Hamilton, Leskovec, and
  Jurafsky}]{hamilton2016diachronic}
William~L. Hamilton, Jure Leskovec, and Dan Jurafsky. 2016{\natexlab{b}}.
\newblock \href {https://doi.org/10.18653/v1/P16-1141} {Diachronic word
  embeddings reveal statistical laws of semantic change}.
\newblock In \emph{Proceedings of the 54th Annual Meeting of the Association
  for Computational Linguistics (Volume 1: Long Papers)}, pages 1489--1501,
  Berlin, Germany. Association for Computational Linguistics.

\bibitem[{Hofmann et~al.(2020)Hofmann, Pierrehumbert, and
  Sch{\"u}tze}]{hofmann2020dynamic}
Valentin Hofmann, Janet~B Pierrehumbert, and Hinrich Sch{\"u}tze. 2020.
\newblock Dynamic contextualized word embeddings.
\newblock \emph{arXiv preprint arXiv:2010.12684}.

\bibitem[{Huang et~al.(2022)Huang, Shao, Chang, Xiong, and
  Hwu}]{huang2022understanding}
Jie Huang, Hanyin Shao, Kevin Chen-Chuan Chang, Jinjun Xiong, and Wen-mei Hwu.
  2022.
\newblock Understanding jargon: Combining extraction and generation for
  definition modeling.
\newblock In \emph{Proceedings of the 2022 Conference on Empirical Methods in
  Natural Language Processing}, pages 3994--4004.

\bibitem[{Jang et~al.(2022)Jang, Ye, Lee, Yang, Shin, Han, Kim, and
  Seo}]{jang2022temporalwiki}
Joel Jang, Seonghyeon Ye, Changho Lee, Sohee Yang, Joongbo Shin, Janghoon Han,
  Gyeonghun Kim, and Minjoon Seo. 2022.
\newblock Temporalwiki: A lifelong benchmark for training and evaluating
  ever-evolving language models.
\newblock \emph{arXiv preprint arXiv:2204.14211}.

\bibitem[{Joshi et~al.(2020)Joshi, Lee, Luan, and
  Toutanova}]{joshi2020contextualized}
Mandar Joshi, Kenton Lee, Yi~Luan, and Kristina Toutanova. 2020.
\newblock Contextualized representations using textual encyclopedic knowledge.
\newblock \emph{arXiv preprint arXiv:2004.12006}.

\bibitem[{Lazaridou et~al.(2021)Lazaridou, Kuncoro, Gribovskaya, Agrawal,
  Liska, Terzi, Gimenez, de~Masson~d'Autume, Kocisky, Ruder
  et~al.}]{lazaridou2021mind}
Angeliki Lazaridou, Adhi Kuncoro, Elena Gribovskaya, Devang Agrawal, Adam
  Liska, Tayfun Terzi, Mai Gimenez, Cyprien de~Masson~d'Autume, Tomas Kocisky,
  Sebastian Ruder, et~al. 2021.
\newblock Mind the gap: Assessing temporal generalization in neural language
  models.
\newblock \emph{Advances in Neural Information Processing Systems}, 34.

\bibitem[{Liu et~al.(2019)Liu, Ott, Goyal, Du, Joshi, Chen, Levy, Lewis,
  Zettlemoyer, and Stoyanov}]{liu2019roberta}
Yinhan Liu, Myle Ott, Naman Goyal, Jingfei Du, Mandar Joshi, Danqi Chen, Omer
  Levy, Mike Lewis, Luke Zettlemoyer, and Veselin Stoyanov. 2019.
\newblock \href {http://arxiv.org/abs/1907.11692} {Roberta: A robustly
  optimized bert pretraining approach}.

\bibitem[{Loureiro et~al.(2022)Loureiro, Barbieri, Neves, Anke, and
  Camacho-Collados}]{loureiro2022timelms}
Daniel Loureiro, Francesco Barbieri, Leonardo Neves, Luis~Espinosa Anke, and
  Jose Camacho-Collados. 2022.
\newblock Timelms: Diachronic language models from twitter.
\newblock \emph{arXiv preprint arXiv:2202.03829}.

\bibitem[{Luu et~al.(2021)Luu, Khashabi, Gururangan, Mandyam, and
  Smith}]{luu2021time}
Kelvin Luu, Daniel Khashabi, Suchin Gururangan, Karishma Mandyam, and Noah~A
  Smith. 2021.
\newblock Time waits for no one! analysis and challenges of temporal
  misalignment.
\newblock \emph{arXiv preprint arXiv:2111.07408}.

\bibitem[{Mickus et~al.(2019)Mickus, Paperno, and Constant}]{mi:19}
Timothee Mickus, Denis Paperno, and Matthieu Constant. 2019.
\newblock \href {https://aclanthology.org/W19-6201} {Mark my word: A
  sequence-to-sequence approach to definition modeling}.
\newblock In \emph{Proceedings of the First NLPL Workshop on Deep Learning for
  Natural Language Processing}, pages 1--11, Turku, Finland. Link{\"o}ping
  University Electronic Press.

\bibitem[{Mickus et~al.(2022)Mickus, Van~Deemter, Constant, and
  Paperno}]{mickus2022semeval}
Timothee Mickus, Kees Van~Deemter, Mathieu Constant, and Denis Paperno. 2022.
\newblock Semeval-2022 task 1: Codwoe--comparing dictionaries and word
  embeddings.
\newblock \emph{arXiv preprint arXiv:2205.13858}.

\bibitem[{Navigli and Velardi(2010)}]{navigli2010learning}
Roberto Navigli and Paola Velardi. 2010.
\newblock Learning word-class lattices for definition and hypernym extraction.
\newblock In \emph{Proceedings of the 48th annual meeting of the association
  for computational linguistics}, pages 1318--1327.

\bibitem[{Osborne et~al.(2014)Osborne, Lall, and
  Van~Durme}]{osborne2014exponential}
Miles Osborne, Ashwin Lall, and Benjamin Van~Durme. 2014.
\newblock Exponential reservoir sampling for streaming language models.
\newblock In \emph{Proceedings of the 52nd Annual Meeting of the Association
  for Computational Linguistics (Volume 2: Short Papers)}, pages 687--692.

\bibitem[{Pilehvar and Camacho-Collados(2019)}]{pilehvar2019wic}
Mohammad~Taher Pilehvar and Jose Camacho-Collados. 2019.
\newblock \href {https://doi.org/10.18653/v1/N19-1128} {{W}i{C}: the
  word-in-context dataset for evaluating context-sensitive meaning
  representations}.
\newblock In \emph{Proceedings of the 2019 Conference of the North {A}merican
  Chapter of the Association for Computational Linguistics: Human Language
  Technologies, Volume 1 (Long and Short Papers)}, pages 1267--1273,
  Minneapolis, Minnesota. Association for Computational Linguistics.

\bibitem[{Raffel et~al.(2020)Raffel, Shazeer, Roberts, Lee, Narang, Matena,
  Zhou, Li, and Liu}]{raffel2020exploring}
Colin Raffel, Noam Shazeer, Adam Roberts, Katherine Lee, Sharan Narang, Michael
  Matena, Yanqi Zhou, Wei Li, and Peter~J Liu. 2020.
\newblock Exploring the limits of transfer learning with a unified text-to-text
  transformer.
\newblock \emph{The Journal of Machine Learning Research}, 21(1):5485--5551.

\bibitem[{Rosin and Radinsky(2022)}]{rosin2022temporal}
Guy~D Rosin and Kira Radinsky. 2022.
\newblock Temporal attention for language models.
\newblock \emph{arXiv preprint arXiv:2202.02093}.

\bibitem[{Rudolph and Blei(2018)}]{rudolph2018dynamic}
Maja Rudolph and David Blei. 2018.
\newblock Dynamic embeddings for language evolution.
\newblock In \emph{Proceedings of the 2018 World Wide Web Conference}, pages
  1003--1011.

\bibitem[{Rudolph et~al.(2016)Rudolph, Ruiz, Mandt, and
  Blei}]{rudolph2016exponential}
Maja Rudolph, Francisco Ruiz, Stephan Mandt, and David Blei. 2016.
\newblock Exponential family embeddings.
\newblock \emph{Advances in Neural Information Processing Systems}, 29.

\bibitem[{Sanh et~al.(2020)Sanh, Debut, Chaumond, and
  Wolf}]{sanh2020distilbert}
Victor Sanh, Lysandre Debut, Julien Chaumond, and Thomas Wolf. 2020.
\newblock \href {http://arxiv.org/abs/1910.01108} {Distilbert, a distilled
  version of bert: smaller, faster, cheaper and lighter}.

\bibitem[{Spala et~al.(2020)Spala, Miller, Dernoncourt, and
  Dockhorn}]{spala2020semeval}
Sasha Spala, Nicholas Miller, Franck Dernoncourt, and Carl Dockhorn. 2020.
\newblock Semeval-2020 task 6: Definition extraction from free text with the
  deft corpus.
\newblock In \emph{Proceedings of the Fourteenth Workshop on Semantic
  Evaluation}, pages 336--345.

\bibitem[{Spala et~al.(2019)Spala, Miller, Yang, Dernoncourt, and
  Dockhorn}]{spala2019deft}
Sasha Spala, Nicholas~A Miller, Yiming Yang, Franck Dernoncourt, and Carl
  Dockhorn. 2019.
\newblock Deft: A corpus for definition extraction in free-and semi-structured
  text.
\newblock In \emph{Proceedings of the 13th Linguistic Annotation Workshop},
  pages 124--131.

\bibitem[{Turc et~al.(2019)Turc, Chang, Lee, and
  Toutanova}]{DBLP:journals/corr/abs-1908-08962}
Iulia Turc, Ming{-}Wei Chang, Kenton Lee, and Kristina Toutanova. 2019.
\newblock \href {http://arxiv.org/abs/1908.08962} {Well-read students learn
  better: The impact of student initialization on knowledge distillation}.
\newblock \emph{CoRR}, abs/1908.08962.

\bibitem[{Vaswani et~al.(2017)Vaswani, Shazeer, Parmar, Uszkoreit, Jones,
  Gomez, Kaiser, and Polosukhin}]{vaswani2017attention}
Ashish Vaswani, Noam Shazeer, Niki Parmar, Jakob Uszkoreit, Llion Jones,
  Aidan~N Gomez, {\L}ukasz Kaiser, and Illia Polosukhin. 2017.
\newblock Attention is all you need.
\newblock \emph{Advances in neural information processing systems}, 30.

\bibitem[{Veyseh et~al.(2020)Veyseh, Dernoncourt, Dou, and
  Nguyen}]{veyseh2020joint}
Amir Veyseh, Franck Dernoncourt, Dejing Dou, and Thien Nguyen. 2020.
\newblock A joint model for definition extraction with syntactic connection and
  semantic consistency.
\newblock In \emph{Proceedings of the AAAI Conference on Artificial
  Intelligence}, volume~34, pages 9098--9105.

\bibitem[{Xu et~al.(2022)Xu, Chen, Liu, Wen, and Yuan}]{xutaxoprompt}
Hongyuan Xu, Yunong Chen, Zichen Liu, Yanlong Wen, and Xiaojie Yuan. 2022.
\newblock Taxoprompt: A prompt-based generation method with taxonomic context
  for self-supervised taxonomy expansion.
\newblock In \emph{Proceedings of the IJCAI Conference on Artificial
  Intelligence}.

\bibitem[{Yarowsky(1995)}]{yarowsky1995unsupervised}
David Yarowsky. 1995.
\newblock Unsupervised word sense disambiguation rivaling supervised methods.
\newblock In \emph{33rd annual meeting of the association for computational
  linguistics}, pages 189--196.

\bibitem[{Yogatama et~al.(2014)Yogatama, Wang, Routledge, Smith, and
  Xing}]{yogatama2014dynamic}
Dani Yogatama, Chong Wang, Bryan~R Routledge, Noah~A Smith, and Eric~P Xing.
  2014.
\newblock Dynamic language models for streaming text.
\newblock \emph{Transactions of the Association for Computational Linguistics},
  2:181--192.

\bibitem[{Yu et~al.(2021)Yu, Zhu, Fang, Yu, Wang, Xu, Zeng, and
  Jiang}]{yu2021dict}
Wenhao Yu, Chenguang Zhu, Yuwei Fang, Donghan Yu, Shuohang Wang, Yichong Xu,
  Michael Zeng, and Meng Jiang. 2021.
\newblock Dict-bert: Enhancing language model pre-training with dictionary.
\newblock \emph{arXiv preprint arXiv:2110.06490}.

\bibitem[{Zervakis et~al.(2022)Zervakis, Vincent, Couceiro, Schoenauer, and
  Marquer}]{zervakis}
Georgios Zervakis, Emmanuel Vincent, Miguel Couceiro, Marc Schoenauer, and
  Esteban Marquer. 2022.
\newblock \href {https://inria.hal.science/hal-03792071} {{An analogy based
  approach for solving target sense verification}}.
\newblock In \emph{{NLPIR 2022 - 6th International Conference on Natural
  Language Processing and Information Retrieval}}, Bangkok, Thailand.

\bibitem[{Zhu et~al.(2020)Zhu, Rawat, Zaheer, Bhojanapalli, Li, Yu, and
  Kumar}]{zhu2020modifying}
Chen Zhu, Ankit~Singh Rawat, Manzil Zaheer, Srinadh Bhojanapalli, Daliang Li,
  Felix Yu, and Sanjiv Kumar. 2020.
\newblock Modifying memories in transformer models.
\newblock \emph{arXiv preprint arXiv:2012.00363}.

\bibitem[{Zhu et~al.(2019)Zhu, Noraset, Liu, Jiang, and Downey}]{zh:19}
Ruimin Zhu, Thanapon Noraset, Alisa Liu, Wenxin Jiang, and Doug Downey. 2019.
\newblock Multi-sense definition modeling using word sense decompositions.

\end{thebibliography}

%\appendix
%\section{Appendix A: Dataset Details and examples}
%\label{sec:appa}
%Various details about the dataset are enlisted here. We also include some examples from our seed dataset. Additionally, we show an example of the prompt that was used to extract the annotations from GPT3.5. It is worth noting that we use 1-shot prompting to extract the annotations.
%\section{Appendix B: Experiment Details}
%\label{sec:appb}
%The various details of experiments and its related configurations are discussed here. There is no hyperparameter optimisation objective in the scope of our experiments, so we provide the standard set of hyperparamters. We use a learning rate of 2e-5, with AdamW optimizer \citep{kingma2017adam} with a weight decay of 0.01 and batch size of 16. We set gradient accumulation steps to 8, for a faster training and set up warm-up steps to 100. We train model for 3 epochs.

\end{document}

\maketitle
\section{Annotation Prompt}

We show in Figure \ref{fig:prompt} (in the next page due to space constraints), for illustrative purposes, one of the four instructions we use for prompting ChatGPT to annotate \textsc{WikiTiDe}. 

\begin{figure}[!t]
    \begin{center}
    \resizebox{\textwidth}{!}{%
    \begin{minipage}{\textwidth}
    \centering
    \fbox{
    \begin{minipage}{\linewidth} %0.75\textwidth}
    \fontsize{10}{11}\selectfont
    
    \textbf{Prompt:} Given the following two $<$\texttt{timespan},\texttt{definition}$>$ pairs for the term \texttt{\{term\}}:
    \begin{enumerate}
        \item \{\texttt{timespan1}\}:\{\texttt{def1}\}
        \item \{\texttt{timespan2}\}:\{\texttt{def2}\}
    \end{enumerate}

    You must return a Python dictionary with the following key-value pairs:

    \begin{itemize}[itemsep=0pt, parsep=0pt, topsep=0pt]
        \item 'label': 1, 2 or 3
            \begin{itemize} [itemsep=0pt, parsep=0pt, topsep=0pt]
                \begin{tabular}[t]{@{}p{\linewidth}@{}}
                    \item 2 if {def1} and {def2} are different AND something fundamental happened to {term} or the knowledge we had about {term} fundamentally changed between the dates \{\texttt{timespan1}\} and \{\texttt{def1}\}.
                    \item 1 if \{\texttt{def1}\} and \{\texttt{def2}\} are different BUT the difference is mostly semantic or aesthetic, not about new or updated knowledge about {\{term\}}.
                    \item 0 if \{\texttt{def1}\} and \{\texttt{def2}\} are conveying basically the same information.
                \end{tabular}
            \end{itemize}
        \item 'confidence': an int between 0 and 10
            \begin{itemize}[itemsep=0pt, parsep=0pt, topsep=0pt]
                \begin{tabular}[t]{@{}p{\linewidth}@{}}
                    \item this score determines how confident you are in your assessment
                \end{tabular}
            \end{itemize}
        \item 'explanation': a string
            \begin{itemize}[itemsep=0pt, parsep=0pt, topsep=0pt]
                \begin{tabular}[t]{@{}p{\linewidth}@{}}
                    \item a short explanation of why you annotated the data with that particular label
                \end{tabular}
            \end{itemize}
    \end{itemize}
      
    Here are some examples:
    \begin{itemize}[itemsep=0pt, parsep=0pt, topsep=0pt]
        \item INPUT:
            \begin{enumerate}
                \item \textit{2009-09-26T18:48:00Z} : \texttt{Muammar Abu Minyar al-Gaddafi} also known simply as Colonel Gaddafi; born 1942) is a a Libyan politician, revolutionary, and political theorist who has been the de facto leader of Libya since a coup in 1969.
                \item \textit{2023-04-01T05:30:00Z} : \texttt{Muammar Muhammad Abu Minyar al-Gaddafi} (c.1942 – 20 October 2011), also known as Colonel Gaddafi, was a Libyan politician, revolutionary, and political theorist.
            \end{enumerate}
        \item OUTPUT:
        
        \texttt{\{\{ 'label': 2,}
        \item[]
        \texttt{\hspace*{2em} 'confidence': 9,}
        \item[]
        \texttt{\hspace*{2em} 'explanation': Gadaffi died somewhere between 2009-09 and 2023-04 and that is reflected in the different tense used in these definitions.} 
        \texttt{\}\}}
            
    \end{itemize}   

    \begin{itemize}[itemsep=0pt, parsep=0pt, topsep=0pt]
        \item INPUT:
            \begin{enumerate}
                \item \textit{2009-09-26T18:48:00Z} : \texttt{"(Every Time I Turn Around) Back in Love Again"} was a hit song for R\&B/funk band L.T.D.
                \item \textit{2019-10-02T14:44:59Z} : \texttt{(Every Time I Turn Around) Back in Love Again"} is a hit song written by Len Ron Hanks and Zane Grey for R\&B/funk band L.
            \end{enumerate}
        \item OUTPUT:
        
            \texttt{\{\{ 'label': 1,}
            \item[]
            \texttt{\hspace*{2em}  'confidence': 8,}
            \item[]
            \texttt{\hspace*{2em} 'explanation': The hit status of the song is different in terms of syntax,in 2009, it is said as the song was a hit song, and in 2019, it is said that it is a hit song, so the verb form is changed, nothing else.} 
            \texttt{\}\}}
            
    \end{itemize}   

    \begin{itemize}[itemsep=0pt, parsep=0pt, topsep=0pt]
        \item INPUT:
            \begin{enumerate}
                \item \textit{2004-04-27T02:34:00Z} : \texttt{Coffee} as a drink is prepared from the Coffee plant.
                \item \textit{2023-03-22T12:59:00Z} : \texttt{Coffee} is a beverage prepared from roasted coffee beans.
            \end{enumerate}
        \item OUTPUT:
        
        \texttt{\{\{ 'label': 0,}
        \item[]
        \texttt{\hspace*{2em} 'confidence': 7,}
        \item[]
        \texttt{\hspace*{2em} 'explanation': 'Both definitions are conveying the same core information about how coffee is prepared.'} 
        \texttt{\}\}}
            
    \end{itemize}   
    \end{minipage}}
    \end{minipage}
    }
    \end{center}
    
    \captionsetup{justification=centering}
    \caption{Instruction to prompt for the task of annotating definition pairs in \textsc{WikiTiDe}.}
    \label{fig:prompt}

\end{figure}

%\section{Implementation Details}
%\label{sec:appb}
%The various details of experiments and its related configurations are discussed here. There is no hyperparameter optimisation objective in the scope of our experiments, so we provide the standard set of hyperparamters. We use a learning rate of 2e-5, with AdamW optimizer \citep{kingma2017adam} with a weight decay of 0.01 and batch size of 16. We set gradient accumulation steps to 8, for a faster training and set up warm-up steps to 100. We train model for 3 epochs.